\def\year{2019}\relax 
\documentclass[letterpaper]{article}  
\usepackage{aaai19}  
\usepackage{times}  
\usepackage{helvet}  
\usepackage{courier}  
\usepackage{url}
\usepackage{graphicx}  
\usepackage{amsfonts}

\usepackage[utf8]{inputenc}

\usepackage{amsmath}   
\usepackage{amsthm}

\usepackage[margin=1in]{geometry}
\usepackage[makeroom]{cancel}
\usepackage{mathtools}

\newcommand{\R}{\mathbb{R}}

\newcommand{\citet}[1]{\citeauthor{#1} (\citeyear{#1})}

\title{MAD-TN: A Tool for Measuring Fluency in Human-Robot Collaboration}
\author{Seth Isaacson\thanks{Authors listed alphabetically but contributed equally.}\\
Harvey Mudd College\\
sisaacson@hmc.edu\\
\And
Gretchen Rice\footnotemark[1]\\
Olin College of Engineering\\
grice@olin.edu\\
\And
James C. Boerkoel Jr.\\
Harvey Mudd College\\
boerkoel@hmc.edu\\
}

\nocopyright

\begin{document}

\maketitle

\begin{abstract}
Fluency is an important metric in Human-Robot Interaction (HRI) that describes the coordination with which humans and robots collaborate on a task. 
Fluency is inherently linked to the timing of the task, making temporal constraint networks a promising way to model and measure fluency. 
We show that the Multi-Agent Daisy Temporal Network (MAD-TN) formulation, which expands on an existing concept of daisy-structured networks, is both an effective model of human-robot collaboration and a natural way to measure a number of existing fluency metrics.
The MAD-TN model highlights new metrics that we hypothesize will strongly correlate with human teammates' perception of fluency. 


\end{abstract}

\section{Introduction}

In the field of HRI, developing a fluent, collaborative schedule requires consideration of a range of parameters: abilities and preferences of teammates, temporal constraints within and between agents' tasks, and team fluidity and efficiency. 
An ideal solution to this problem adequately addresses each of these concerns and facilitates intuitive measurement of fluency.

Consider a situation in which a human and a robot are collaborating on a packaging task. 
Each agent may excel at certain tasks, and struggle with or be unable to complete others. 
In the packaging example, the robot may be better suited to retrieve an object from a dangerous machine. 
If the robot is stationary, a human would be much better suited to retrieve an object placed across the room.
Additionally, some tasks may depend on the completion of other tasks: before the robot can deliver the package, the human must finish sealing it.
However, two actions often do not depend on each other, such as retrieving two objects from two different locations.
Ideally, the coordination between agents should be sufficiently fluent as to mimic the experience of human-human collaboration.


In this paper, we amend the Daisy model proposed by \citet{MANIADAKIS0} so that it more generally and accurately captures the activities and temporal constraints of human-robot teams while facilitating measures of fluency. 
The resulting Multi-Agent Daisy Temporal Network (MAD-TN) formalizes daises in the vocabulary of temporal networks. 
We show that this model is well-suited for intuitively representing collaborative, real-world tasks while also enhancing the monitoring and management of team fluency.
Finally, we propose two new metrics that we hypothesize will correlate with human perception of team fluency. We further provide a new lens for understanding an existing fluency metric---functional delay---on two scales.


 

\section{MAD-TN}
We introduce the Multi-Agent Daisy Temporal Network (MAD-TN)---which we also call a Daisy---as a scheduling problem formulation that supports fluent collaboration in multi-agent tasks.
At a high level, a daisy models multi-agent systems by breaking large tasks into smaller tasks that can be completed independently. 
Our work builds on the work of  \citeauthor{MANIADAKIS0}, who first proposed daisy-structured temporal networks for use in the context of multi-agent collaboration (\citeyear{MANIADAKIS0}). 
We modify their formulation to achieve a more general model while maintaining the helpful traits they identified.
Our model more precisely characterizes the daisy framework in the language of temporal networks.
We also formalized actions to have start and end times, which in turn, are useful for better understanding the timing of interactions between agents



\subsection{Background: Temporal Constraint Networks}
A Temporal Constraint Network (TCN) is generally a set $T$ of timepoints where $t_i \in \mathbb{R}$ with a set $C$ of constraints on those timepoints. 
These networks are often encoded as directed graphs where nodes are timepoints and edges represent the constraints. 
An assignments of times to each $t_i \in T$ that satisfies all the constraints is called a \textit{schedule}. A network with at least one valid schedule is called \textit{consistent}.
The Simple Temporal Network (STN) is an example of a TCN with constraints of the form $t_j - t_i \in [-c_{ji}, c_{ij}]$, where $c_{ij},c_{ji} \in \R$ \cite{DECHTER}. 
A Multi-agent STN generalizes STNs by designating which agents are responsible for the scheduling and execution of each timepoint \cite{MASTP}.
Finally, a Disjunctive Temporal Network (DTN) is similar to an STN, but allows for disjunctive constraints of the form $\bigvee_k \left( t_j^k - t_i^k \in [-c_{ji}^k, c_{ij}^k] \right)$, which represent a disjunctive choice among different simple temporal constraint options. 
While the DTN model is a more powerful representation that allows deciding how to order events, finding a solution for a disjunctive network is NP-complete, whereas STNs are polynomial time solvable \cite{TEMPORAL_COMPLEXITY}.

\subsection{Example Human-Robot Packaging Task}
\begin{figure}
  \centering
  \includegraphics[width=200pt]{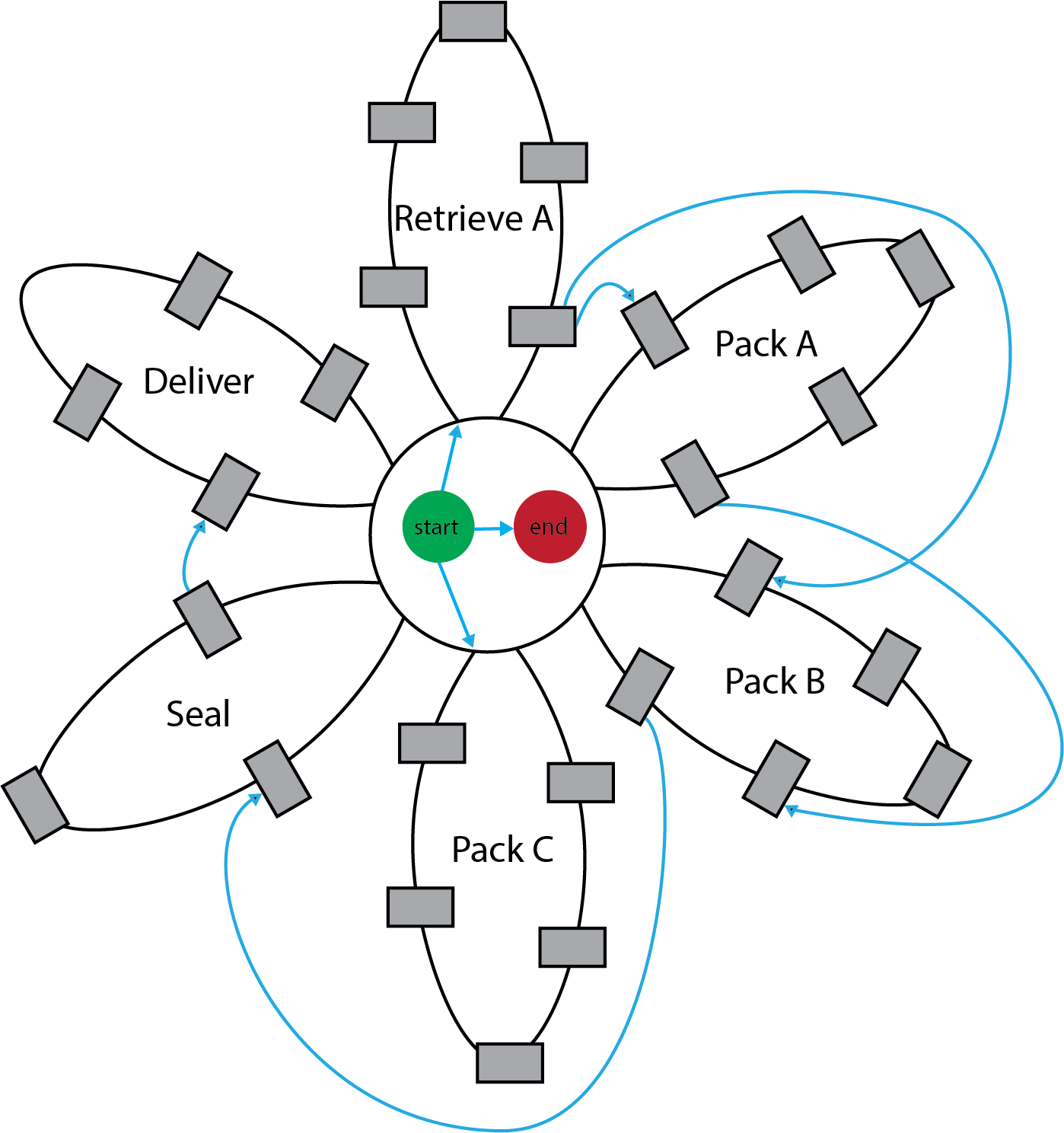}
  \caption{A daisy model of the experiment, adapted from \citet{MANIADAKIS0}. Each petal is labeled as the sub-task it represents and includes minimal actions, represented by the grey boxes. The blue arrows represent external constraints between petals.}
  \label{fig:daisy}
\end{figure}

Next, we describe the components of the daisy and how to measure fluency within it using an example packaging scenario as shown in Figure \ref{fig:daisy}.   
In this packaging scenario, we pair one robot with one human. 
This collaborative packing task contains sub-tasks such as: ``Retrieve Object A", ``Prepare and Pack Object B", ``Pack Object A", ``Pack Object C", ``Seal Package", and ``Deliver Package." 
Each sub-task will be assigned to the human or the robot. 
This task was designed to have distinct hand-off points where a resource (such as Object A) is transferred from one agent to another.
We believe these interactions are important for fluency.

\subsection{Actions}
An action contains a start vertex, end vertex, and a makespan constraint between the two. 
The makespan constraint should always constrain the durations of actions to be non-negative.
To provide the most flexibility and control when scheduling, actions should represent atomic tasks, such as moving to a location or picking up an object. 
In the packaging example, one action in the ``Retrieve Object A" subtask is ``Pick Up Object A." 
This action contains a ``Start Pick" vertex and ``Finish Pick" vertex and might have a constraint between the nodes with bounds [0.5, 3] meaning the pick may take anywhere from 0.5 to 3 seconds.

\subsection{Petals}
\begin{figure}
  \centering
  \includegraphics[width=175pt]{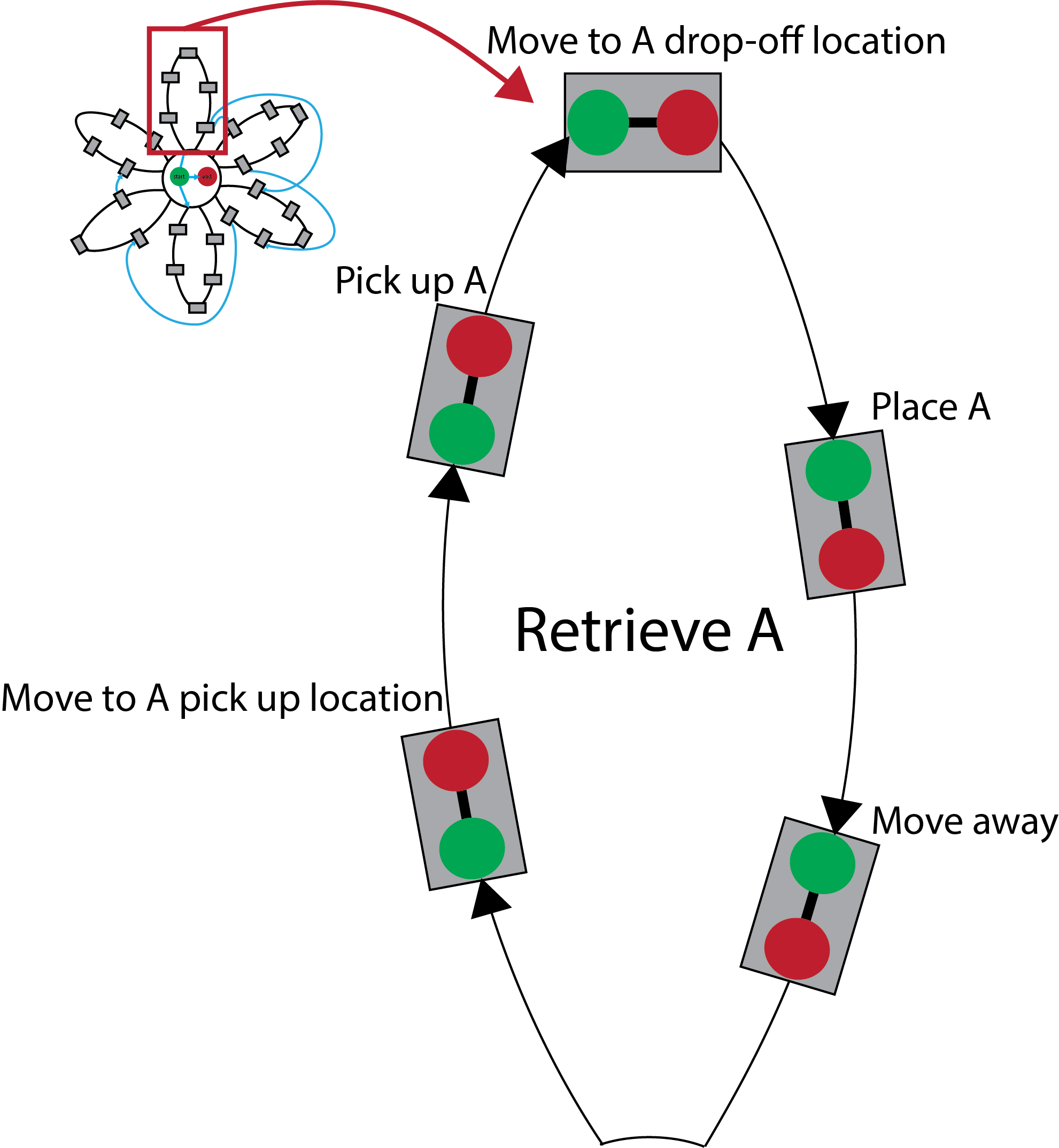}
  \caption{The ``retrieve A" petal containing five actions represented as labeled grey boxes.}
  \label{fig:petal}
\end{figure}

A petal serves to break an overall collaborative task into sub-tasks that can be done by a single agent. 
A petal contains a sequence of actions, connected by a set of constraints that imputes the order in which the actions are completed.
A petal should generally deal with the handling of a single resource or set of resources. 
We assume that transition times between actions within a petal are negligible unless an agent must wait due to an external constraint (explained next). 
Figure \ref{fig:petal} illustrates the ``Retrieve Object A" petal from the packaging task.


\subsection{Daisy Structure}
A daisy consists of a set of petals, and any constraints between them. 
More formally, $D = \langle P , C, V_s, V_e\rangle$ where $P=\{P_1, \cdots, P_k\}$ for a daisy $D$ of $k$ petals. 
$V_s$ is a start vertex representing the start time of the daisy, and $V_e$ is an end vertex representing the end time of the daisy. 
$C$ is the set of constraints that are not contained within one petal.
A constraint in $C$ between $V_s$ and $V_e$ restricts the overall makespan of the daisy.

A daisy's petals may be completed in \emph{any} order, subject to the constraints between them.
For example, since Object A has to be retrieved before it can be packaged, there would be a handoff constraint between the ``Retrieve A" and ``Pack A" petals, as illustrated in Figure \ref{fig:daisy}.
This mandates that the ``Retrieve A" petal cannot follow the ``Pack A" petal.
So while petals can generally be done in any order, $C$ often enforces a partial ordering of petals.
For our discussion, we will assume that a consistent total ordering of petals has already been determined.

Each petal in the daisy is completed by an agent.  
Generally, petals can be assigned to any agent capable of completing the subtask and once assigned, that petal becomes part of their local subproblem.
The way agents are assigned to petals is an interesting planning problem that likely influences the overall fluency of the collaborative task. 
For instance, \citet{DAISYPLAN} developed a greedy planner that assigns petals based on each agents ability. 
However, for the purposes of this discussion, we assume that each petal has been preassigned to either the human or robot.





\section{Fluency}
According to \citeauthor{FLUENCY}, fluency is a ``coordinated meshing of joint activities between members of a well-synchronized team" (\citeyear{FLUENCY}). 
Hoffman discusses human idle time, robot idle time, concurrent activity, and functional delay as quantitative metrics of fluency.
The MAD-TN provides a natural way to measure each of these metrics.
Further, the MAD-TN structure allows us to define two \emph{new} fluency metrics: \emph{concurrent inactivity} and \emph{resource delay}, both of which we hypothesize will impact human teammates' perceptions of fluency.

\subsection{Agent Idle Time}

Agent idle time refers to the amount of time an agent is not executing an action.
In the daisy, an agent's idle time is the sum across two sources of idle time. 
The first is rest time, which is the amount of time an agent is within a daisy but waiting to begin work on a petal (i.e., the time spent resting between subtasks).
The second is time spent between actions.
For most actions, this source of idle time will be negligible, since we assume agents will complete actions within their subtasks/petals without pause.
However, an agent may be forced to wait due to an external constraint involving another petal.  


\subsection{Concurrent Activity}

Concurrent activity is the amount of time both agents are active at the same time.
Concurrent activity is simple to measure in the MAD-TN structure as the sum of the overlapping action times. 

\subsection{Functional Delay}

Functional delay (F-DEL) is the delay (positive or negative) between when one agent stops working on a task and another agent begins work on a dependent task. 
Functional delay can be separated into the delay caused by a transition from a robot to a human or \textit{vice-versa}.

Functional delay can be measured on daisies on two scales---at the petal/sub-task level and at the action level. 
Here we discuss it on the petal scale.
If two petals owned by differing agents share a constraint, a dependence is implied between the petals. 
Take, for example, when the human must seal the box (``Seal" petal in Figure \ref{fig:daisy}) before the robot delivers it (``Deliver" petal in Figure \ref{fig:daisy}). 
The robot's delivery depends on the human's sealing. 
This introduces the possibility of robot functional delay. 
That delay is measured as the difference between when the robot begins work on it's petal minus the time when the human ends work on their petal. 
If that time is positive, it indicates the robot introduced positive functional delay.
Negative functional delay indicates the robot anticipated the human's action and began its petal before the human finished theirs.
We discuss action-level functional delay below.


\subsection{Concurrent Inactivity}

We introduce a new metric, \textbf{concurrent inactivity}, that measures the amount time both agents are simultaneously inactive.
Since the daisy structure naturally allows parallel task execution, agents should mostly always be active. 
Thus, both agents being inactive is a worst-case-scenario for efficiency. 
We expect concurrent inactivity to be inversely correlated with perceived fluency.

\subsection{Resource Delay}
When agents collaborate over a shared resource, a fluent handoff of that resource would be one where one agent finishes its use of the resources just in time for the other agent to begin using it.
Thus, a fluent handoff is characterized by \textbf{Resource Delay (R-DEL)} --- a \emph{new} metric we introduce --- and action-level functional delay. 
Resource delay captures the time between when an agent is ready to use a resource and when that resource becomes available.
Action-level functional delay captures the time between when a resource is available and when an agent begins to use it.

Consider the situation in which the human is moving Object A to an accessible location for the robot to pack.
If the robot arrives before the box has been placed down, it is being \textit{blocked} by the human. 
This would be a positive resource delay. 
On the other hand, if the robot arrives after the human places Object A, the handoff becomes \textit{stale}, which is a negative resource delay. 
In summary, R-DEL is the amount of time (positive or negative) between when one agent is ready to start an action and when the that action is enabled by the other agent involved in the handoff. 

In our precise definitions of resource and action-level functional delay below, we focus on typical handoffs where one agent must complete the use of one resource prior to yielding it to the other agent.
As such, the definitions that follow assume these external handoff constraints will have a lower bound of 0.
In the future, we hope to extend our definitions of functional delay and resource delay to other types of external constraints, including e.g., synchronization constraints or handoffs with positive wait times associated them.

\begin{figure}
  \centering
  \includegraphics[width=220px]{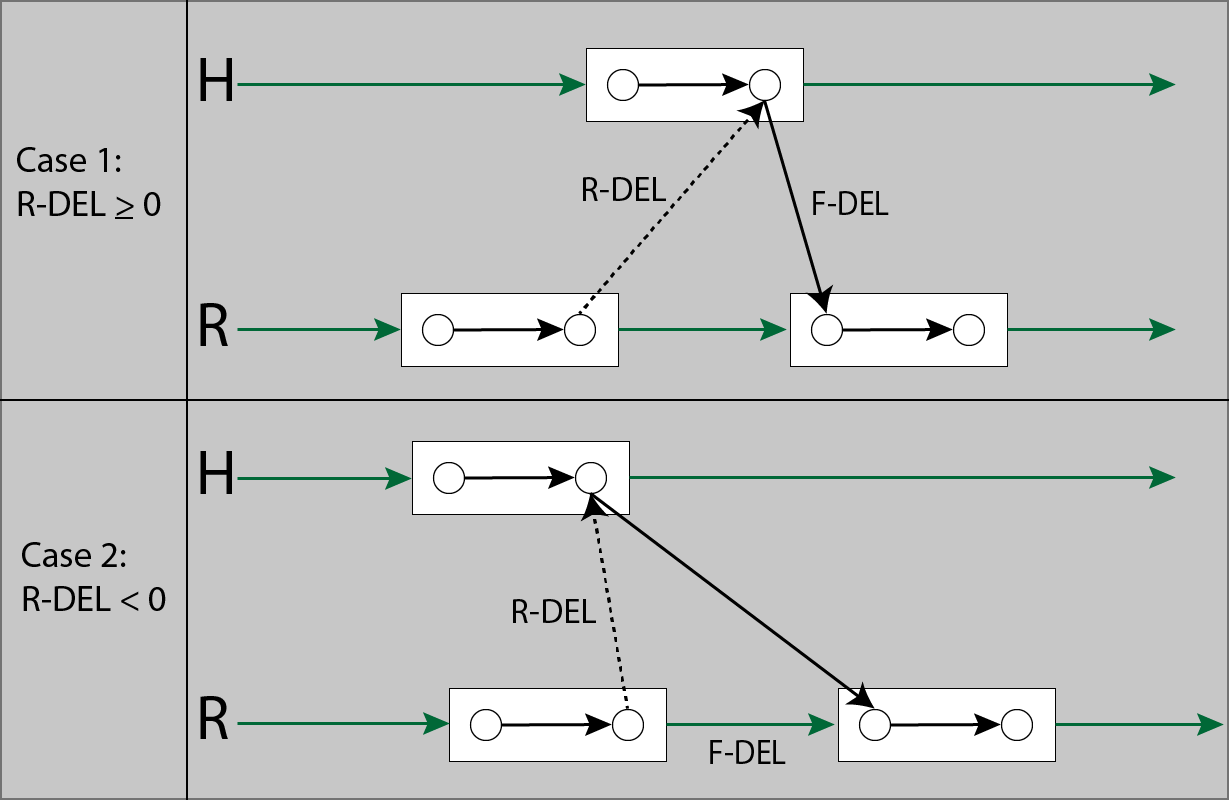}
\caption{Resource delay (dashed line) and functional delay (solid line) for a scenario where the human blocks robot progress (Case 1) and one where the resource becomes stale (Case 2). Here, white boxes are actions while the contained circles are start and end nodes.}
  \label{fig:r-del}
\end{figure}

When there is a positive resource delay, action-level functional delay is the delay between when one agent enables the others' action and the time that agent begins, e.g., the delay between the human placing Object A and the robot beginning to pack Object A.
A positive functional delay represents the transition time between agents due to the handoff, e.g., the time it takes the robot to process that Object A has been placed as shown as Case 1 in Figure \ref{fig:r-del}.
A negative functional delay represents anticipatory action, e.g., the robot began picking up object A before the human finished placing it down.

In the second case of a negative resource delay, we instead define functional delay to be the time it takes the agent to process and act on the fact that the resource is already available, and thus is measured as the time between the start of the activity and the end of the preceding one.
In our running example, this would be the time between when the robot finishes the ``Move to Object A" action and when it begins the ``Pick Object A" action. 
Since the resource delay was stale, functional delay in this situation can only be positive (since the agent has arrived late, it is now impossible to begin work early). 
A situation with staleness and positive resource delay can be seen in Figure \ref{fig:r-del} in Case 2.

\subsection{Human Perceptions of Fluency}

\citet{FLUENCY} validated each of the existing fluency metrics against human perceptions of fluency using a simulated human-robot interaction in which a human and robot completed alternating tasks involving manipulating objects in a common workspace.
He used an online platform to have participants watch videos of the simulation and answer questions about how fluent the interactions appeared.
He found a statistically significant correlation with human idle time and the viewers perception of fluency, but not with robot idle time or concurrent activity.
The strongest correlation reported was between function delay and fluency. 

Overall, we believe the MAD-TN will provide a useful tool for empirically exploring the efficacy of these five metrics at predicting human's perceptions of fluency.
Indeed, our ongoing investigation includes plans to design, implement, and analyze a physically embodied human-robot experiment that tests correlations of each of these with human's perception.
We posit that the degree to which these fluency metrics correlate with human perceptions of team fluency may be influenced by the experimental setup.
For instance, a human's teammate perception of fluency might be much more impacted by an idle physically-embodied robot teammate than an observer of a simulation.
Further, we believe that the concurrent nature of MAD-TNs will highlight the importance of some metrics such as concurrent activity and concurrent inactivity.
Finally, we are particularly motivated to further investigate the role that resource delay plays in human teammates' perceptions of fluency. 
We suspect a human whose progress is impeded by their robot teammate might cause undue frustration or dissatisfaction.
We believe our ongoing explorations will be key in understanding, designing, and planning more fluent human-robot interactions.

\section{Conclusion}
In this paper we introduced the Multi-Agent Daisy Temporal Network, or MAD-TN, which generalizes an existing model \cite{MANIADAKIS0} by more precisely characterizing it in the language of temporal networks.
 The MAD-TN provides a convenient framework for measuring the fluidity of human-robot interactions using both existing measures and our new fluency metrics, Resource Delay and Concurrent Inactivity.

As we continue this work, we plan to further formalize the MAD-TN and empirically assess it's usefulness in supporting fluent human-robot interactions.
We also plan to design an experiment that validates our new and existing fluidity metrics.
In particular, we hypothesize that collaborative tasks with significant resource delay will hurt overall perceptions of fluidity by human teammates.
Eventually, we hope to explore how to order and assign petals to agents to maximize team efficiency and fluency.


While beyond the scope of this paper, we believe there are many interesting extensions of the MAD-TN that would be useful in human-robot teaming.
We believe the disjunctive nature of petals will be a critical feature in scenarios that are more complex than the planned packaging task.
Further, we believe MAD-TN could be extended to accommodate recursively defined daisy networks, where, e.g., actions within petals could themselves be daisies and petals of these sub-daisies could be assigned to sub-teams of agents.
We also believe our action-level resource and functional delay measures will be critical for actions that require explicit synchronization such as when a robot physically hands an object over to a human teammate \cite{handovers}.
Finally, the petal structure supports interesting questions of efficient resource allocation and issues of the privacy and autonomy of agents' tasks.

\newpage

\section{Acknowledgements}
Funding for this work was graciously provided by the National Science Foundation under grants IIS-1651822 and
CNS-1659805.  Thanks to the
anonymous reviewers, HMC faculty, staff and HEATlab
members for their support and constructive feedback.

\bibliographystyle{aaai.bst}
\bibliography{ref}

\end{document}